\pdfoutput=1

\documentclass[11pt]{article}

\usepackage[review]{EMNLP2023}

\usepackage{times}
\usepackage{latexsym}

\usepackage[T1]{fontenc}

\usepackage[utf8]{inputenc}

\usepackage{microtype}

\usepackage{inconsolata}
\usepackage{graphicx}
\usepackage{amsmath}
\usepackage{amsthm}
\usepackage{booktabs}
\usepackage{algorithm}
\usepackage{algorithmic}
\usepackage[switch]{lineno}
\usepackage{amssymb}
\usepackage{booktabs}
\usepackage{multirow}
\usepackage{makecell}
\usepackage{bm}
\usepackage{appendix}
\usepackage{booktabs}
\title{Evaluation and Analysis of Hallucination in Large Vision-Language Models}

\author{Junyang Wang$^{\clubsuit, }$\thanks{\, Equal contribution} , Yiyang Zhou$^{\spadesuit, *}$, Guohai Xu$^{\triangle}$, Pengcheng Shi$^{\spadesuit}$, Chenlin Zhao$^{\Diamond}$,\\\bf  Haiyang Xu$^{\triangle}$, Qinghao Ye$^{\triangle}$, Ming Yan$^{\triangle}$, Ji Zhang$^{\triangle}$, Jihua Zhu$^{\spadesuit}$, Jitao Sang$^{\clubsuit, }$\thanks{\, Corresponding author\newline Work done during internship at DAMO Academy, Alibaba Group.}, Haoyu Tang$^{\heartsuit, \dag}$ \\
$^{\clubsuit}$ School of Computer and Information Technology, Beijing Jiaotong University, Beijing, China \\
$^{\spadesuit}$ School of Software Engineering, Xi’an Jiaotong University, Xi’an, China \\
$^{\heartsuit}$ School of Software, Shandong University, Jinan, China \\
$^{\Diamond}$ MAIS, Institute of Automation, Chinese Academy of Sciences(CASIA), Beijing, China \\
$^{\triangle}$ DAMO Academy, Alibaba Group\\
{
{\{junyangwang,jtsang\}@bjtu.edu.cn}, 
{\{zhouyiyangailab\}@gmail.com},
{\{guohai.xgh, ym119608\}}@alibaba-inc.com
}
}

\begin{document}
\maketitle
\begin{abstract}
Large Vision-Language Models (LVLMs) have recently achieved remarkable success. However, LVLMs are still plagued by the hallucination problem, which limits the practicality in many scenarios. Hallucination refers to the information of LVLMs' responses that does not exist in the visual input, which poses potential risks of substantial consequences. There has been limited work studying hallucination evaluation in LVLMs. In this paper, we propose \textbf{Ha}llucination \textbf{E}valuation based on Large \textbf{L}anguage \textbf{M}odels (HaELM), an LLM-based hallucination evaluation framework. HaELM achieves an approximate 95\% performance comparable to ChatGPT and has additional advantages including low cost, reproducibility, privacy preservation and local deployment. Leveraging the HaELM, we evaluate the hallucination in current LVLMs. Furthermore, we analyze the factors contributing to hallucination in LVLMs and offer helpful suggestions to mitigate the hallucination problem. Our data and code are available at \url{https://github.com/junyangwang0410/HaELM}.
\end{abstract}

\section{Introduction}
The success of Large Language Models (LLMs), with ChatGPT as a prominent example, has attracted widespread attention~\cite{zhang2022opt, chowdhery2022palm, touvron2023llama, scao2022bloom}. Recently, Large Vision-Language Models (LVLMs) extend LLMs to understand visual inputs and demonstrate impressive multi-modal capabilities in a zero-shot manner~\cite{zhu2023minigpt, liu2023visual, ye2023mplug, gong2023multimodal, wang2023visionllm, li2023otter, mu2023embodiedgpt, su2023pandagpt, liu2020visual}. These efforts have driven the development of multi-modal artificial general intelligence.


\begin{figure}[t] 
\centering
\includegraphics[width=0.475\textwidth]{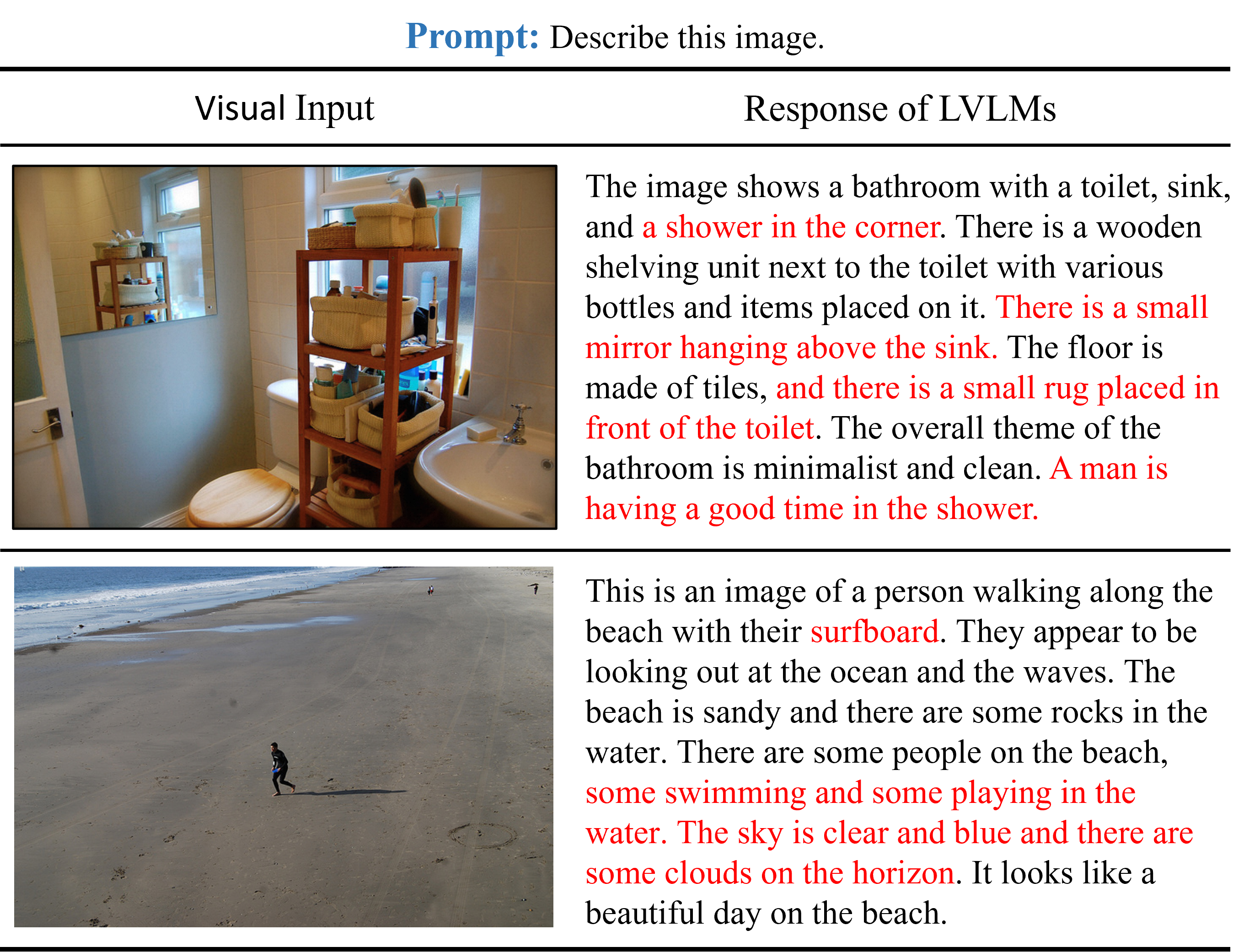}
\caption{Examples of the LVLMs' hallucination. In real-world scenarios, LVLMs may generate content that doesn't match the visual input. The words with red font represent the hallucination.
}
\label{fg1} 
\end{figure}
However, LVLMs still suffer from hallucination which refers to the generation of incorrect information that does not align with the visual input~\cite{liu2023aligning}. Previous work has mainly focused on investigating hallucination in LLMs and Vision-Language Pre-trained Models (VLPMs). For LLMs, hallucination predominantly stems from incorrect knowledge present in the training data~\cite{zhang2023language,li2023halueval}, while for VLPMs, the challenge lies in accurately representing visual information within abstract visual encodings~\cite{shen2021much,biten2022let}. Although LVLMs combine the strengths of both LLMs and VLPMs, they inherently inherit both two pathways of hallucination generation. In this case, the flawed recognition of visual information within the framework of LLMs can lead to deceptively plausible yet ultimately absurd responses, as exemplified in Figure~\ref{fg1}. The hallucination poses potential risks of substantial consequences that need to be addressed and rectified~\cite{li2023evaluating}.

To solve the problem of hallucination in LVLMs, \cite{li2023evaluating} proposed POPE, an object-based hallucination evaluation framework. POPE initially employs an object detector to identify all objects within an image and subsequently utilizes predefined prompts, such as "Is there a \{object\} in this image?", to query the model about the presence of an object which does not exist in the image. The model's response of "yes" is regarded as an indication of hallucination. Nevertheless, our investigation, as shown in Figure~\ref{fg4}, reveals that LVLMs tend to exhibit a response of "yes" to over 80\% of queries about non-existent objects. In contrast, when the prompt "Describe the image" is adopted, less than 10\% of the resultant responses included the hallucination objects. This discrepancy underscores the weak correlation between object-based hallucination evaluation and the actual hallucination of LVLMs.

The above analysis demonstrates that in idealized hallucination evaluation scenarios, LVLMs are highly susceptible to the influence of prompts, leading to biased responses that cannot be used as a basis for hallucination evaluation. Therefore, we advocate for the conduction of hallucination evaluation within real-world scenarios to avoid the negative impact of prompts on the evaluation results. However, one challenge is that the responses of LVLMs in real-world scenarios tend to be complex, which implies that traditional match-based evaluation methods will no longer be applicable. This means that the evaluation tool needs to understand the complex responses of LVLMs. 

We notice that LLMs demonstrate powerful text-understanding capabilities. Based on this, we propose an innovative framework called \textbf{Ha}llucination \textbf{E}valuation based on Large \textbf{L}anguage \textbf{M}odels (HaELM). First, we identify the hallucination patterns exhibited by LVLMs and systematically collect their hallucination responses. Subsequently, we craft prompts that elicit responses from ChatGPT aligned with these patterns to collect the pertinent training data. Finally, We fine-tune LLaMA~\cite{touvron2023llama} through the LoRA-based methodology~\cite{hu2021lora}. As a result, HaELM becomes proficient in hallucination evaluation, leveraging reference descriptions of images as a basis for assessment. Experimental results demonstrate attest to the comparable performance of HaELM and ChatGPT, exhibiting alignment with human annotations. In addition, HaELM has additional advantages including low cost, reproducibility, privacy preservation and local deployment. Finally, we conduct a comprehensive analysis of the factors contributing to hallucination generation in current LVLMs, culminating in a set of suggestions for mitigating the hallucination.

We summarize the contributions as follows:
\begin{itemize}
\item Through our analysis, we discover that LVLMs are easily influenced by prompts in idealized hallucination scenarios, making the results not correlated with hallucinations in real-world scenarios.
\item To our knowledge, we are the first to utilize LLM for hallucination evaluation within LVLMs. We propose \textbf{Ha}llucination \textbf{E}valuation based on Large \textbf{L}anguage \textbf{M}odels (HaELM). HaELM achieves a strong performance and has additional advantages including low cost, reproducibility, privacy preservation and local deployment compared to ChatGPT.
\item Leveraging the HaELM, we embark on evaluating the presence of hallucination in current LVLMs. We analyze the factors that affect hallucination and offer helpful suggestions.
\end{itemize}

\section{Background}
In this section, we mainly introduced existing Large Language Models (LLMs) and Large Vision-Language Models (LVLMs), as well as hallucination problems that exist in LLMs and LVLMs.

\subsection{Large Language Model}
GPT-3 \cite{brown2020language} has demonstrated that language models with a large number of parameters possess powerful zero-shot capabilities and are capable of excelling at previously unseen tasks.
Thanks to the success of GPT-3, now LLMs \cite{zhang2022opt, chowdhery2022palm, touvron2023llama, scao2022bloom} have gained significant attention.
To make LLMs more responsive to human instructions, InstructGPT \cite{ouyang2022training} introduced the instruction-following fine-tuning paradigm. It employs reinforcement learning from human feedback to train the LLMs to follow human instructions and produce desired outputs.

\begin{figure}[t] 
\centering
\includegraphics[width=0.475\textwidth]{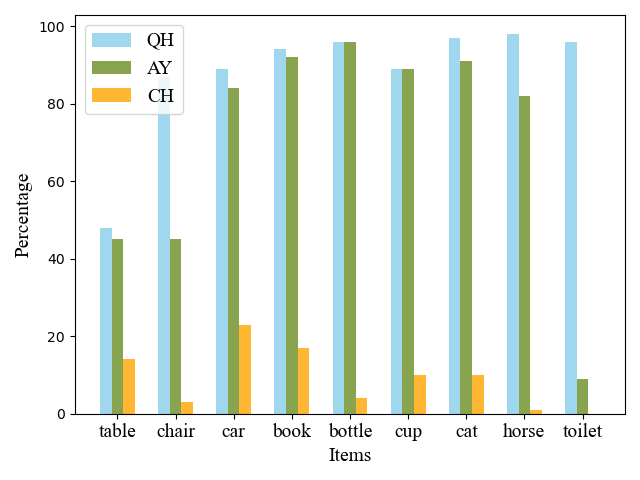}
\caption{The validity assessment results of object-based hallucination evaluation. QH represents the percentage that we asked about the corresponding item on images where it was not present; AY represents the percentage that the model answered "yes", and CH represents the percentage that the model had hallucinations of the corresponding item in the responses.
}
\label{fg4} 
\end{figure}

\subsection{Large Vision-Language Model}
With the success of LLMs, many researchers have been extending language models to understand real-world images.
For example, some approaches \cite{yang2023mm, shen2023hugginggpt} are based on visual expert and regards ChatGPT as the central work.
On the other hand, some recent open-source works such as \cite{zhu2023minigpt, liu2023visual, ye2023mplug, gong2023multimodal, wang2023visionllm, li2023otter, mu2023embodiedgpt, su2023pandagpt} achieve unified LVLMs by aligning extracted visual tokens from a visual encoder with a pre-trained LLM and instruct tuning it. To further improve the performance of LVLMs, \cite{liu2023aligning,li2023m3it} proposed to increase the diversity of instructions and construct the larger instruction fine-tuning dataset.

\subsection{Hallucinations in LLMs and LVLMs}
The issue of hallucinations has been extensively studied in the traditional field of NLP. Despite the advancements in the latest and widely acclaimed LLMs, they remain encumbered by the persistent challenge of hallucinations. Consequently, a multitude of works have emerged, aiming to mitigate the impact of these hallucinations. However, it is noteworthy that limited focus has been directed toward addressing the hallucination in LVLMs \cite{zhou2023analyzing, liu2023aligning}. 

In contrast to hallucinations observed in LLMs, hallucinations within LVLMs arise from a mismatch between the visual and textual modalities. Currently, the only work that specifically focuses on the hallucination of LVLMs utilizing object detection and query instructions \cite{li2023evaluating}. Through meticulous empirical experiments, they substantiate the considerable severity of hallucinations in LVLMs, particularly in generating objects that are absent from the provided images but appear frequently in the training data. The existing LLMs, by adopting instruct tuning, make their target outputs follow human instructions, but this can result in biased training and target distributions \cite{tian2023just}. Furthermore, insufficient visual constraints contribute to the serious issue of illusions in LVLMs.

The presence of hallucinations can lead to unreliability in models, which may cause harm to human society, such as the misleading information output by the model leading to errors in human decision-making or the output of toxic information.

\begin{figure*}[t]
\centering
\includegraphics[width=0.85\textwidth]{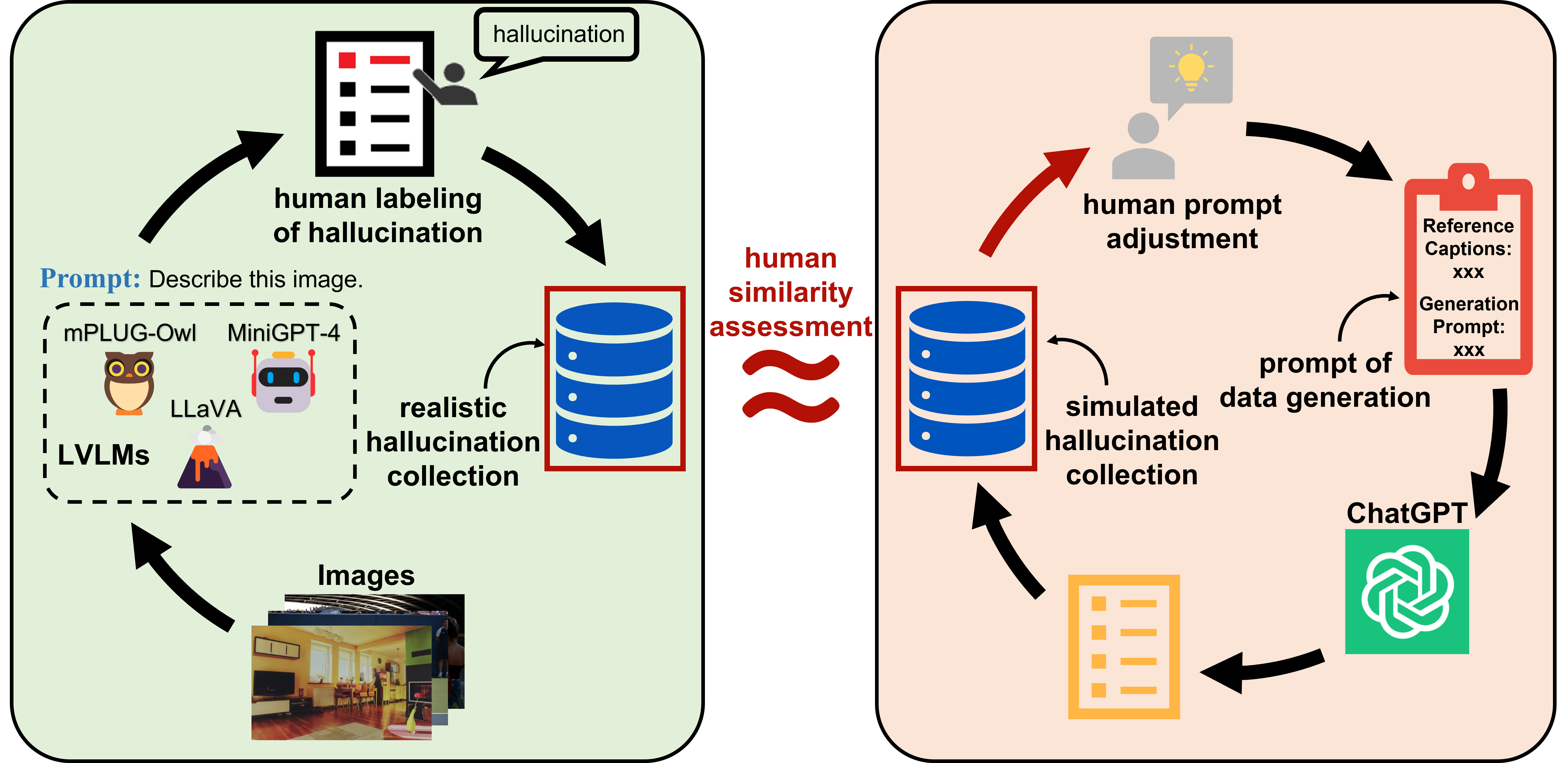}
\caption{The illustration for data collection process of HaELM. 
The left figure illustrates the process of manually collecting real hallucination responses, while the right figure illustrates the generation of data in bulk using ChatGPT. The human similarity assessment aims to align the patterns of simulated hallucination data with realistic one.}
\label{fg2} 
\end{figure*}

\section{Motivation}

The current existing method for hallucination evaluation is object-based hallucination evaluation~\cite{li2023evaluating}. It measures the extent of hallucination in LVLMs by querying their response to the presence of an "item". The "item" is chosen from a list of commonly hallucinated words that do not exist in the image. If the model believes that an item is present in an image where it is absent, it indicates that the model has a hallucination regarding that item.

To verify the feasibility, we designed an experiment based on the object-based hallucination evaluation method. We utilized the prompt "Is there a \{item\} in this photo?" to query mPLUG-Owl regarding 100 randomly selected images from the MS-COCO 2014 dataset~\cite{lin2014microsoft,chen2015microsoft}. Other models' and detailed results are provided in the appendix. The \{item\} in the prompt was substituted with the top ten most frequently hallucinated words proposed by~\cite{li2023evaluating} that are not present in the given image. The results are presented in Figure~\ref{fg4}. The "QH" and "AY" reveal that LVLMs answer "yes" to over 80\% of the queries in this prompt, even if all the items in the prompts were absent from the image.

The above phenomenon can be explained by the tendency of LVLMs to affirm the description when answering judgment-type queries with a "yes" response. 
We speculate that this bias is due to the instruction fine-tuning data that includes a substantial number of responses catering to human requests, which results in bias in LVLMs' responses to judgment-type queries. To verify the relationship between the responses of LVLMs to such queries and corresponding hallucinations, we conducted a manual evaluation in real-world scenarios. We used the prompt "Describe this image" and examined whether the generated descriptions truly contained hallucinations for the items that received a "yes" response. The "AY" and "CH" in Figure~\ref{fg4} reveal that only 10\% of the responses included hallucinations for specific items. This suggests that the hallucinations measured object-based evaluation merely exploit the judgment bias present in LVLMs, rather than reflecting their hallucination.

\section{Method}
This section mainly introduces the definition of hallucination and our method of Hallucination Evaluation based on Large Language Models.

\begin{figure*}[t] 
\centering
\includegraphics[width=0.95\textwidth]{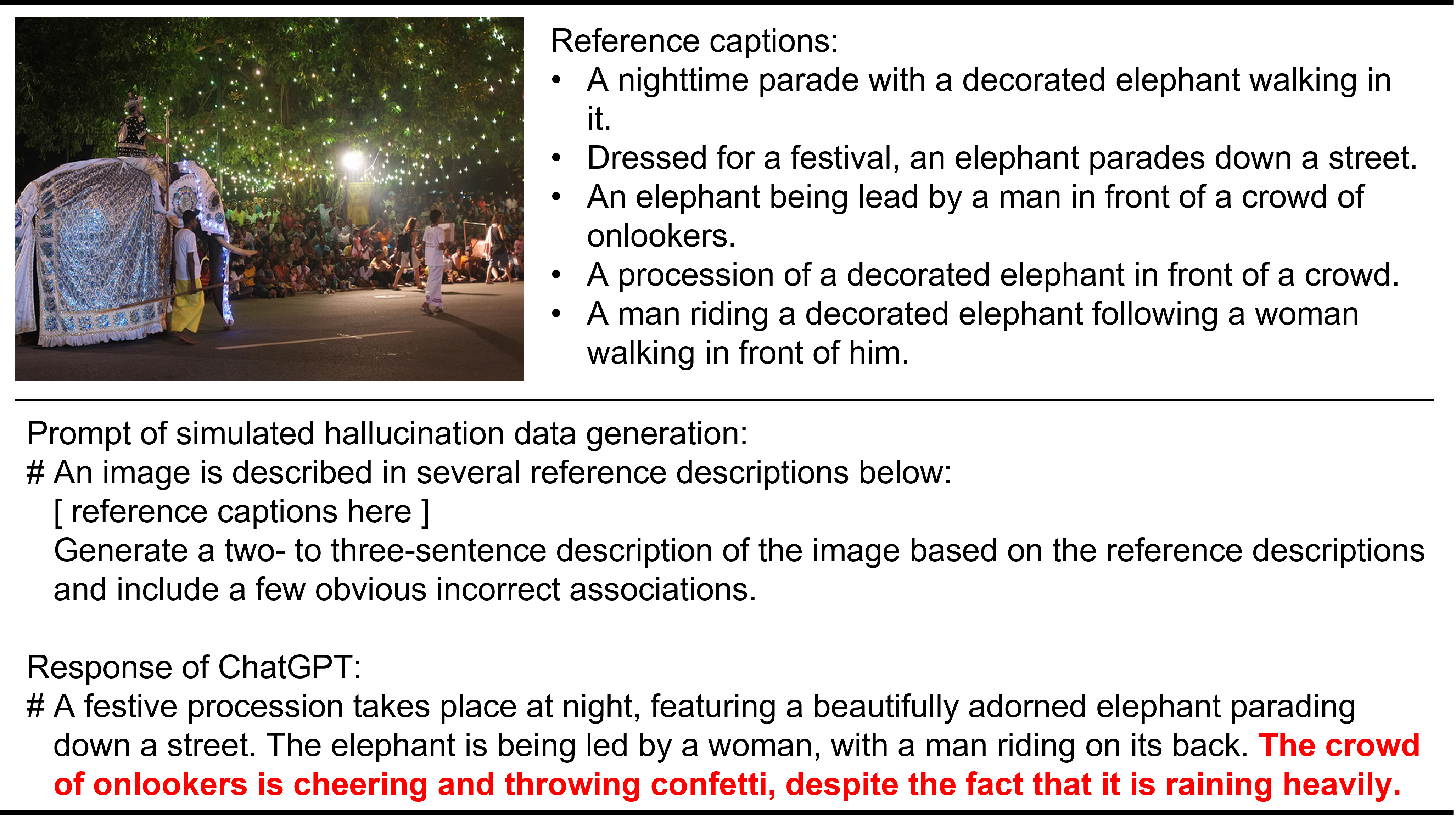}
\caption{An example of the prompt for generating simulated hallucination samples, where the words with red font represent the hallucination description.}
\label{fg3} 
\end{figure*}

\subsection{Problem Definition}

The evaluation of hallucinations in real-world scenarios for LVLMs is defined as determining whether there are discrepancies between the content of the images and the responses generated by LVLMs, under the potential requests that could be initiated by humans. In this paper, we focus on the real-world scenario of image description.

\subsection{HaELM}
\paragraph{Data Collection}~\\
To perceive hallucinations in the responses of LVLMs, it is crucial to evaluation on both non-hallucinatory and hallucinatory responses. To address this, we first analyze the hallucination patterns of LVLMs. Randomly selecting images, we query the LVLMs with the instruction "Describe this image" and manually annotated the hallucination responses to get the realistic hallucination collection as shown in the left of Figure~\ref{fg2}.

Subsequently, our goal is to obtain a substantial amount of hallucination data in bulk. We considered using ChatGPT to generate hallucinations by manually constructing prompts based on the reference captions of the images provided. We compared the hallucination data generated by ChatGPT with realistic hallucination data by human similarity assessment. We iteratively modified the prompt to make the patterns of the two align closely as shown in the right of Figure~\ref{fg2}. Our hallucination data collection format is presented in Figure~\ref{fg3}.

Finally, we collect the non-hallucination data. By requesting ChatGPT to generate detailed descriptions based on reference captions, we can easily obtain the desired data. However, it is crucial to emphasize that the generated descriptions should strictly adhere to the objects present in the reference captions, without introducing any non-existent elements.

\paragraph{Training and Inference}~\\
During the training phase, we employ a consistent format prompt that corresponds to the data distribution of LLMs and instruction fine-tuning. The collected data from the preceding step is seamlessly integrated into the prompt, serving as the training data for fine-tuning the LLM through an autoregressive training process. During the inference phase, we incorporate the reference captions and responses from the LVLMs under evaluation into the prompt. These inputs are then fed into the meticulously trained evaluation model to get the judgment.

HaELM can be reused multiple times once data collection and training finish, which offers a cost advantage over ChatGPT while ensuring reproducibility. Furthermore, HaELM is built upon an open-source LLM, allowing for local deployment, thereby eliminating uploading data and guaranteeing data privacy.

\section{Experiments}
\paragraph{Dataset}~\\
Our image dataset consists exclusively of images from the MS-COCO 2014~\cite{lin2014microsoft,chen2015microsoft}, following the established partition into the train, val and test sets as outlined by~\cite{karpathy2015deep}. For data collection purposes, we randomly select 10,000 samples from the training set and collect 10,000 hallucination and 10,000 non-hallucination simulated responses respectively. Additionally, we obtain all 5,000 samples from the test set specifically for evaluating the LVLMs' hallucinations. To ensure consistency and accuracy in our data collection and hallucination evaluation, we use the manually annotated captions provided in the dataset as reference captions.

\begin{table*}[t]
    \centering
    \renewcommand{\arraystretch}{1.2}
    \setlength{\tabcolsep}{7.5pt}
    \scalebox{0.90}{
    \begin{tabular}{l c c c c c c c c c c c c}
    \hline
    \multirow{2}{*}{\textbf{Method}}&\multicolumn{4}{c}{\textit{w/o hallucination}}&\multicolumn{4}{c}{\textit{w/ hallucination}}&\multicolumn{4}{c}{\textit{all}}\\
    \cmidrule(lr){2-5}
            \cmidrule(lr){6-9}
            \cmidrule(lr){10-13}
    &LL&Mi&mP&Avg.&LL&Mi&mP&Avg.&LL&Mi&mP&Avg.\\
            \hline
    GPT-3.5&82.0&38.9&50.8&57.2&\textbf{48.7}&\textbf{78.1}&\textbf{72.9}&\textbf{66.6}&\textbf{69.0}&\textbf{64.0}&\textbf{59.0}&\textbf{64.0}\\
            HaELM&\textbf{93.4}&\textbf{61.1}&\textbf{60.1}&\textbf{71.5}&25.6&57.8&43.2&42.2&67.0&59.0&57.0&61.0\\
            \hline
      \end{tabular}
    }
    \caption{The results of accuracy on human-annotated evaluation data for HaELM and GPT-3.5, where LL, Mi, and mP respectively represent LLaVA, Mini-GPT4, and mPLUG-Owl.}
    \label{tab:human_result1}
\end{table*}

\begin{table*}[t]
	\centering  
	\renewcommand{\arraystretch}{1.2}
	\setlength{\tabcolsep}{6pt}
	\scalebox{0.90}{
	\begin{tabular}{l c c c c c c c c c c c c}
		\hline
		\multirow{2}{*}{\textbf{Method}}&\multicolumn{3}{c}{LLaVA}&\multicolumn{3}{c}{MiniGPT-4}&\multicolumn{3}{c}{mPLUG-Owl}\\
		\cmidrule(lr){2-4}
            \cmidrule(lr){5-7}
            \cmidrule(lr){8-10}
		&Precision&Recall&F1 Score&Precision&Recall&F1 Score&Precision&Recall&F1 Score\\
            \hline
		&\multicolumn{9}{c}{\textit{w/o hallucination}}\\
            \hline
		GPT-3.5&\textbf{71.4}&82.0&76.3&\textbf{50.0}&38.9&43.8&\textbf{76.2}&50.8&61.0\\
            HaELM&66.3&\textbf{93.4}&\textbf{77.5}&44.9&\textbf{61.1}&\textbf{51.8}&66.1&\textbf{65.1}&\textbf{65.6}\\
            \hline
            &\multicolumn{9}{c}{\textit{w/ hallucination}}\\
            \hline
            GPT-3.5&63.3&\textbf{48.7}&\textbf{55.0}&69.4&\textbf{78.1}&\textbf{73.5}&\textbf{46.6}&\textbf{73.0}&\textbf{56.8}\\
            HaELM&\textbf{71.4}&25.6&37.7&\textbf{72.5}&57.8&64.3&42.1&43.2&42.7\\
            \hline
            &\multicolumn{9}{c}{\textit{average}}\\
            \hline
            GPT-3.5&67.4&\textbf{65.4}&\textbf{65.6}&\textbf{59.7}&58.5&\textbf{58.7}&\textbf{61.4}&\textbf{61.9}&\textbf{58.9}\\
            HaELM&\textbf{68.9}&59.5&57.6&58.7&\textbf{59.5}&58.1&54.1&54.2&51.7\\
            \hline
    	\end{tabular}
	}
    \caption{The results of accuracy on human-annotated evaluation data for HaELM and GPT-3.5 in terms of precision, recall, and F1 score for hallucination and non-hallucination responses.}
	\label{tb:human_result2}
\end{table*}

\paragraph{Implementation Details}~\\
We employed the LLaMA~\cite{touvron2023llama} as a foundation model and utilized LoRA~\cite{hu2021lora} for fine-tuning. Our hyperparameter is presented in Table~\ref{parameter} of appendix. The training process required 2 hours using a single Tesla V100 GPU. For the evaluated models, we selected the currently available open-source LVLMs: mPLUG-Owl~\cite{ye2023mplug}, MiniGPT-4~\cite{zhu2023minigpt} and LLaVA~\cite{liu2023visual}. The parameter settings are presented in Table~\ref{model} of appendix. We chose the state-of-the-art LLM, ChatGPT, as our baseline.

To ensure the model's focus on hallucination evaluation, we disabled gradient computations on the input, preventing the learning of irrelevant information. Furthermore, our training data outputs were explicitly limited to "yes" or "no" responses, effectively benefiting the automated evaluation.

When evaluating hallucinations by ChatGPT, we further enhanced the accuracy through manual prompt editing, ensuring a fair basis for comparison. Notably, we refrained from employing manually annotated real hallucination data in the training process to uphold the integrity and reliability of our experimental findings.

\subsection{Evaluation on HaELM}
In this subsection, we first evaluate the performance of HaELM. As we are the first to utilize LLM for hallucination evaluation, we select the highly competitive ChatGPT as our baseline for comparative analysis. Given the absence of an established benchmark, we use the realistic hallucination responses derived from LVLMs during the data collection phase as the evaluation benchmark and the annotations as the ground truth.

\paragraph{Accuracy}~\\
We first compared the accuracy. The experimental results on human-annotated hallucination, non-hallucination and overall responses are summarized in Table~\ref{tab:human_result1}. Notably, HaELM achieves an accuracy of 61\%, slightly lower than ChatGPT's performance at 64\%. Nevertheless, HaELM demonstrates an impressive capability, reaching 95\% of ChatGPT's level. 

We also noticed that HaELM performs better in non-hallucination responses, while ChatGPT performs better in hallucination responses. This reflects the biases in the decision-making of the two methods. ChatGPT tends to believe that responses have hallucinations, while HaELM leans towards non-hallucination responses. We analyzed that although simulated hallucination responses mostly cover the hallucination pattern, they still cannot fully match the distribution of actual hallucination responses. Therefore, HaELM fails to learn some patterns of hallucinations, resulting in misclassification under these patterns.

\paragraph{Refined Metrics}~\\
We then proceeded to evaluate the refined metrics, including precision, recall, and F1 scores as shown in Table~\ref{tb:human_result2}. The average F1 scores reveal that HaELM achieves performance levels of 88\%, 99\%, and 88\% on the three LVLMs, respectively. Additionally, as mentioned in the previous analysis, the recall for hallucination responses is lower for HaELM. Nevertheless, despite this limitation, HaELM outperforms ChatGPT in several metrics.

\paragraph{Time \& Cost}~\\
HaELM only requires one-time data collection and training for reuse, allowing significant time and cost savings in subsequent evaluation processes compared to ChatGPT. We present the cost comparison between the two in Table~\ref{cost}.

\begin{table}[!ht]
	\centering
	\renewcommand{\arraystretch}{1.2}
	\setlength{\tabcolsep}{5pt}
	\scalebox{0.90}{
	\begin{tabular}{l c c c c c c}
    \hline
    \multirow{2}{*}{\textbf{Method}}&\multicolumn{2}{c}{Collection}&\multicolumn{2}{c}{Training}&\multicolumn{2}{c}{*Evaluation}\\
    \cmidrule(lr){2-3}
    \cmidrule(lr){4-5}
    \cmidrule(lr){6-7}
    &Time&Cost&Time&Cost&Time&Cost\\
    \hline
    GPT3.5&-&-&-&-&1.6h&6.6\$\\
    HaELM&1.8h&4.3\$&2h&-&0.2h&-\\
    \hline
	\end{tabular}
	}
    \caption{The time and cost of hallucination evaluation for HaELM and ChatGPT. *Evaluation represents a single evaluation conducted on three LVLMs.}
	\label{cost}
\end{table}

HaELM requires only 3.8 hours and 4.3\$ for data collection and training, resulting in a saving of 1.4 hours and 6.6\$ per evaluation compared to ChatGPT. This advantage becomes more significant when multiple evaluations are needed, such as exploring the impact of prompts on hallucinations. Additionally, HaELM can be deployed locally, eliminating the need for internet connectivity and ensuring data and privacy protection.

\subsection{Evaluation on Hallucination}
In this subsection, we will employ HaELM to evaluate the hallucination performance of existing LVLMs. Additionally, we explore the correlation between various generation settings and hallucinations in LVLMs, thereby presenting viable suggestions to mitigate hallucinations.

\paragraph{Comparison on LVLMs}~\\
We evaluate the hallucination of LVLMs across various prompts for a generation. The experimental results are shown in Table~\ref{tab:ha_eva}. Firstly, it can be seen that among these three LVLMs, LLaVA exhibits the lowest degree of hallucination and sensitivity to prompts, far below the other two models. However, previous work~\cite{ye2023mplug} manually annotated results indicate that LLaVA performs the worst in various aspects. This observation aligns with our understanding of LVLMs. We note that the generation of hallucination is often positively correlated with the model's generative capability. For example, hallucinations are almost impossible to occur in VLPMs. Therefore, there exists a trade-off between model performance and hallucinations, which deserves researchers to invest more effort in model selection.

\begin{table}[ht]
    \centering  
	\renewcommand{\arraystretch}{1.2}
	\setlength{\tabcolsep}{6pt}
	\scalebox{0.85}{
	\begin{tabular}{l | c c c c | c}
    \hline
    Model&P1&P2&P3&P4&Avg-M\\
    \hline
    LLaVA&20.0&19.4&18.6&19.5&19.4\\
    MiniGPT-4&46.1&35.5&69.7&68.8&55.0\\
    mPLUG-Owl&35.9&24.1&47.2&37.6&36.2\\
    \hline
    Avg-P&34.0&26.3&45.2&42.0&-\\
    \hline
    \end{tabular}
	}
    \caption{Hallucination evaluation results for LVLMs. The numbers represent the frequency of hallucinations exhibited by the respective LVLM when using generation prompts on the MS-COCO 2014 test split. "Avg-M" represents the average hallucination ratio of the corresponding model across multiple prompts, while "Avg-P" represents the average hallucination ratio of the corresponding prompt across multiple models.\\
    P1: "Describe this image."\\
    P2: "Generate a caption for this image."\\
    P3: "Please restore the scene in the image with words."\\
    P4: "What is this?"}
    \label{tab:ha_eva}
\end{table}

Secondly, it can be observed that both MiniGPT-4 and mPLUG-Owl suffer from severe hallucination issues. The performance of these two models is highly dependent on the choice of prompts. This means that prompt selection should be careful when using these powerful LVLMs.

\begin{figure*}[t] 
\centering
\includegraphics[width=0.8\textwidth]{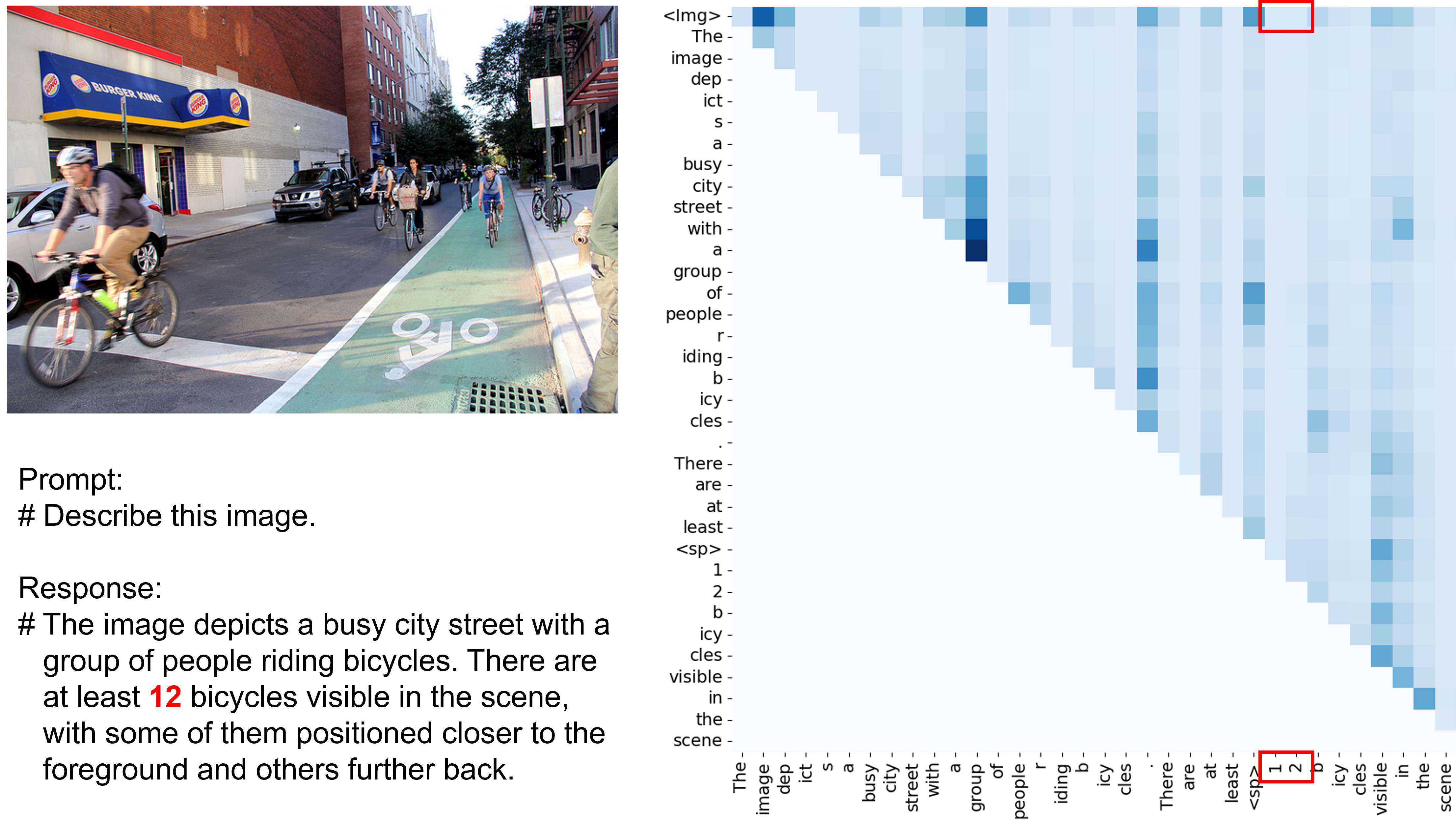}
\caption{We visualized the attention of LVLM during the autoregressive generation. In the right figure, the horizontal axis represents the tokens to be generated, and the vertical axis represents the tokens that have already been generated. "<Img>" represents the average attention on the image, and "<sp>" represents the token "space".}
\label{fg5} 
\end{figure*}

\paragraph{Comparison on Generation Length}~\\
We noticed that in Table~\ref{tab:ha_eva}, using the prompt "Generate a caption for this image." resulted in a minimal amount of hallucination. We collected responses from LVLMs under this prompt and observed that these responses were relatively shorter and more concise. We hypothesize that the generation length of LVLMs' responses may be related to hallucination. To validate this idea, we conducted experiments with mPLUG-Owl by selecting different maximum generation lengths and using the prompt "Describe this image." for a generation. The experimental results are shown in Table~\ref{length}.

\begin{table}[!ht]
	\centering
	\renewcommand{\arraystretch}{1.2}
	\setlength{\tabcolsep}{8pt}
	\scalebox{0.90}{
	\begin{tabular}{l c c c c}
    \hline
    max length&128&256&512&1024\\
    \hline
    hallucination&33.1&35.7&35.9&37.0\\
    \hline
	\end{tabular}
	}
    \caption{The result of comparison on generation length.}
	\label{length}
\end{table}

We observed that as the maximum length increased, the hallucination became stronger. We manually collected a portion of responses with a maximum generation length of 1024 and found that hallucinations tended to occur more toward the latter part of the responses. In this pattern of hallucination, LVLMs often generated a concise segment first, followed by a divergence of imagination. However, this is not always the case, as the examples shown in Figure~\ref{fg1} also demonstrated that LVLMs can generate hallucinations in the earlier parts. Therefore, this represents only a trend. We suggest that obtaining relatively accurate results can be achieved by truncating the responses.

\paragraph{Comparison on Sampling}~\\
Sampling can control LVLMs to generate diverse responses. The current mainstream sampling method is top-$K$ sampling, which randomly selects from the top $K$ words with the highest probabilities each time. To investigate the impact of sampling methods on illusions, we controlled the value of $K$ in top-$K$ sampling and conducted experiments. The experimental results are presented in Table~\ref{topk}.

\begin{table}[!ht]
	\centering
	\renewcommand{\arraystretch}{1.2}
	\setlength{\tabcolsep}{6pt}
	\scalebox{0.90}{
	\begin{tabular}{l c c c c c}
    \hline
    K&1&2&3&4&5\\
    \hline
    hallucination&24.7&33.0&35.9&40.3&42.4\\
    \hline
	\end{tabular}
	}
    \caption{The result of comparison on $K$ of sampling.}
	\label{topk}
\end{table}

Clearly, as $K$ increases, the hallucination issue becomes more severe. Random sampling may cause LVLMs to choose tokens that are less aligned with the visual input, resulting in factual errors. These errors can be rationalized under LLMs, ultimately forming hallucinations. There is still a trade-off between diversity and hallucination.

\section{Discussion}

A comprehensive understanding of the causes behind hallucination in LVLMs remains elusive, as no previous work has been able to provide a definitive explanation. In this section, we aim to shed light on this phenomenon by delving into an analysis of attention using specific visualization techniques. 

We leverage gradients to visualize the attention of each token generated concerning the previously generated tokens and the image. Specifically, we begin by disabling random sampling to ensure the stability of model generation and record the model's generated response. Subsequently, we utilize this response as a label for gradient back-propagation, ultimately obtaining gradients concerning the input embeddings. Finally, we normalize the gradient variations to obtain attention. In Figure~\ref{fg5}, we show an example of hallucination.

We observe that during the occurrence of the hallucination "12", the model exhibits minimal attention to the image (highlighted by the red box). Additionally, the attention of token "1" is primarily focused on the preceding token "<sp>", and the attention of token "2" is also not concentrated in relevant regions. It is possible that tokens "<sp>" and "1" appeared frequently during the training phase, leading the model to learn a biased false correlation. This inherent bias in the LVLM causes the attention during the generation of certain tokens to deviate from the image.

This finding is insightful and carries significant implications. It demonstrates that one possible approach to addressing hallucinations could be to penalize attention that deviates from the image. This will be further explored in our future work.

\section{Conclusion}

In this paper, we analyzed the problems within the existing hallucination evaluation method and proposed HaELM, a hallucination evaluation framework based on LLM designed for real-world scenarios. We demonstrated through experiments that HaELM achieves performance comparable to that of ChatGPT. Building upon HaELM, we conducted analyses on the causes of hallucinations and provided corresponding suggestions to mitigate them. Additionally, our visualization results may hold insightful implications for future research.

\section{Limitations}

Firstly, both HaELM and ChatGPT fall short of achieving human-level hallucination evaluation performance. We attribute this to the fact that current methods are based on language models, using reference captions as a substitute for images. This means that the evaluation models cannot truly comprehend the content of the images. Moreover, we have also attempted to use multimodal models for evaluation. Unfortunately, current LVLMs commonly exhibit hallucinations themselves. Therefore, at this stage, language models remain the optimal choice for hallucination evaluation.

Secondly, we did not address the root cause of hallucinations in LVLMs. In this paper, we investigated the triggers of hallucination and based on this, substantive methods should be established through the analysis of these triggers to reduce the model's learning of hallucination patterns during the training phase. Currently, this is a challenging task for us, but it will remain one of our future work.

\bibliographystyle{acl_natbib}
\bibliography{custom}

\section*{Appendix}
\appendix

\section{Evaluated LVLMs}

We present detailed parameter settings of the evaluated LVLMs, as shown in Table~\ref{model}.

\begin{table}[!ht]
	\centering
	\renewcommand{\arraystretch}{1.2}
	\setlength{\tabcolsep}{5pt}
	\scalebox{0.88}{
	\begin{tabular}{l c c c}
    \hline
    \textbf{Model}&VE&AN&LLM\\
    \hline
    mPLUG-Owl&ViT-L/14&Attention&LLaMA-7B\\
    MiniGPT-4&ViT-G/14&Linear&Vicuna-13B\\
    LLaVA&ViT-L/14&Linear&LLaMA-13B\\
    \hline
	\end{tabular}
	}
    \caption{The detailed parameter settings of the evaluated LVLMs, where VE, AN, LLM stand for Visual Encoder, Alignment Network and Large Language Model, respectively.}
	\label{model}
\end{table}

\begin{table}[!ht]
	\centering
	\renewcommand{\arraystretch}{1.2}
	\setlength{\tabcolsep}{8pt}
	\scalebox{0.95}{
	\begin{tabular}{c c}
    \hline
    base model&LLaMA-7B\\
    batch size&64\\
    epoch&3\\
    learning rate&3e-4\\
    max input length&512\\
    LoRA r&8\\
    LoRA alpha&16\\
    LoRA dropout&0.05\\
    LoRA module&Q\&V\\
    train on input&False\\
    train with fp16&True\\
    \hline
	\end{tabular}
	}
    \caption{The detailed parameter settings.}
	\label{parameter}
\end{table}

\begin{table*}[!ht]
    \centering
    \renewcommand{\arraystretch}{1.2}
    \setlength{\tabcolsep}{8pt}
    \scalebox{0.90}{
    \begin{tabular}{l | c c c c c c c c c c | c}
    \hline
    Item&person&table&chair&car&book&bottle&cup&cat&horse&toilet&sum\\
    \hline
    QH&48&87&89&94&96&89&97&98&96&96&890\\
    AY&45&45&84&92&96&89&91&82&9&84&717\\
    CH&14&3&23&17&4&10&10&1&0&0&82\\
    \hline
    \end{tabular}
    }
    \caption{The detailed validity assessment results of object-based hallucination evaluation method by mPLUG-Owl.}
    \label{tab:mp}
\end{table*}

\begin{table*}[!ht]
    \centering
    \renewcommand{\arraystretch}{1.2}
    \setlength{\tabcolsep}{8pt}
    \scalebox{0.90}{
    \begin{tabular}{l | c c c c c c c c c c | c}
    \hline
    Item&person&table&chair&car&book&bottle&cup&cat&horse&toilet&sum\\
    \hline
    QH&48&87&89&94&96&89&97&98&96&96&890\\
    AY&22&49&51&58&49&44&47&45&21&46&432\\
    CH&6&7&13&10&2&0&3&3&0&1&46\\
    \hline
    \end{tabular}
    }
    \caption{The detailed validity assessment results of object-based hallucination evaluation method by MiniGPT-4.}
    \label{tab:mi}
\end{table*}

\begin{table*}[!ht]
    \centering
    \renewcommand{\arraystretch}{1.2}
    \setlength{\tabcolsep}{8pt}
    \scalebox{0.90}{
    \begin{tabular}{l | c c c c c c c c c c | c}
    \hline
    Item&person&table&chair&car&book&bottle&cup&cat&horse&toilet&sum\\
    \hline
    QH&48&87&89&94&96&89&97&98&96&96&890\\
    AY&42&49&83&91&95&82&94&92&38&87&753\\
    CH&8&2&16&9&2&4&8&0&0&0&49\\
    \hline
    \end{tabular}
    }
    \caption{The detailed validity assessment results of object-based hallucination evaluation method by LLaVA.}
    \label{tab:ll}
\end{table*}

\section{Training Details}

We present detailed parameter settings of the LoRA fine-tuning during the training phase, as shown in Table~\ref{parameter}.

Due to the insufficient 32GB memory of the Tesla V100 to accommodate a batch size of 64, we used a batch size of 8 with a gradient accumulation of 8 steps to achieve an equivalent batch size of 64. When "train on input" is turned off, the self-regressive loss will no longer consider the input part. In addition, fp16 can accelerate training with almost no impact, so we chose to enable it. We adopted the settings from Vicuna for LoRA and replaced the weights of the Q and V matrices.

\section{Additional Evaluation on Hallucination}

The temperature in LLMs generation parameters refers to the parameter that controls the randomness of language model generation during text generation. It is a parameter that controls randomness and can influence the diversity and creativity of model generation to a certain extent.

In principle, the temperature parameter recalibrates the probability distribution of model output, making the probability distribution more evenly distributed. In high-temperature conditions, more probabilities are assigned to lower probabilities, so the generated text is more diverse. In low-temperature conditions, more probabilities are assigned to high-probability results, so the generated text tends to have common patterns.

We conducted experiments to investigate whether the diversity brought by high temperatures would enhance the generation of hallucinations. The results are shown in Table~\ref{temperture}. It can be seen from the results that the hallucinations of the model are enhanced with the increase in temperature, which is consistent with our intuitive judgment. The enhancement of diversity may lead to the generation of unreasonable texts, which are likely to be part of hallucinations. Therefore, we recommend considering a low temperature if the authenticity of the generated texts needs to be ensured.

\begin{table}[!ht]
	\centering
	\renewcommand{\arraystretch}{1.2}
	\setlength{\tabcolsep}{6pt}
	\scalebox{0.92}{
	\begin{tabular}{l c c c c c}
    \hline
    temperture&0.2&0.4&0.6&0.8&1\\
    \hline
    hallucination&24.7&26.6&31.1&33.0&35.9\\
    \hline
	\end{tabular}
	}
    \caption{The result of comparison on temperture.}
	\label{temperture}
\end{table}

\section{Detailed Results}

We present detailed results of the object-based hallucination evaluation. mPLUG-OWl, MiniGPT-4, and LLaVA are shown in Table~\ref{tab:mp}, Table~\ref{tab:mi}, and Table~\ref{tab:ll}, respectively. In the table, QH represents the number of times we asked about the corresponding item on images where it was not present; AY represents the number of times the model answered "yes", and CH represents the number of times the model had hallucinations of the corresponding item in the generated captions.

We observed that the conclusions obtained from the main text apply to almost all LVLMs, indicating that the limitations of object-based hallucination evaluation are not accidental. We realized that LVLMs are highly susceptible to prompt induction in artificially constructed ideal hallucination scenarios.

\end{document}